\documentclass[conference]{IEEEtran}
\IEEEoverridecommandlockouts
\usepackage{cite}
\usepackage{amsmath,amssymb,amsfonts}
\usepackage{algorithmic}
\usepackage{graphicx}
\usepackage{textcomp}
\usepackage{xcolor}
\usepackage[utf8]{inputenc}
\usepackage[utf8]{inputenc}
\DeclareUnicodeCharacter{2013}{--}
\usepackage{url}
\def\BibTeX{{\rm B\kern-.05em{\sc i\kern-.025em b}\kern-.08em
    T\kern-.1667em\lower.7ex\hbox{E}\kern-.125emX}}
\begin{document}

\title{Robust Probabilistic Load Forecasting for a Single Household: A Comparative Study from SARIMA to Transformers on the REFIT Dataset\\
}

\author{\IEEEauthorblockN{ Manoj, Midhun}
\IEEEauthorblockA{
\textit{Senior Platform Engineer, Envestnet}\\
Midhun.Manoj@envestnet.com}

}

\maketitle

\begin{abstract}
Probabilistic forecasting is essential for modern risk management, allowing decision-makers to quantify uncertainty in critical systems. This paper tackles this challenge using the volatile REFIT household dataset, which is complicated by a large structural data gap. We first address this by conducting a rigorous comparative experiment to select a Seasonal Imputation method, demonstrating its superiority over linear interpolation in preserving the data's underlying distribution. We then systematically evaluate a hierarchy of models, progressing from classical baselines(Seasonal naïve, SARIMA) to machine learning and deep learning architectures .Our findings reveal that classical models fail to capture the data's non-linear, regime-switching behaviour. While the LSTM achieved the best-calibrated probabilistic forecast, the Temporal Fusion Transformer (TFT) emerged as the superior all-round model, achieving the best point forecast accuracy (RMSE 481.94) and producing safer, more cautious prediction intervals that effectively capture extreme volatility.
\end{abstract}
\footnote{All project code and analysis notebooks are available at: \url{https://github.com/middhun-31/Robust-Probabilistic-Load-Forecasting-for-a-Single-Household}}

\section{\textbf{Introduction}}
Risk handling plus choices in fine-grained weather guesses need something beyond just a single fixed number. Unlike predictions stuck on one exact outcome, which give only a lone result as if it’s guaranteed, chance-based forecasts show how unsure things are by laying out whole ranges of what might happen. That matters most when used in money-related tasks, moving goods around, or power use planning - where knowing the odds of rare situations usually counts way more than focusing on normal ones.Guessing electricity demand soon ahead for individual homes stands as a classic example - one that's been around long but still pushes limits in daily practice. Knowing the full range - not just averages - of power use helps companies keep grids steady, boost response efforts during peak times while handling more unpredictable wind and solar output. \\Home electricity patterns jump around, shift suddenly, often depend on personal habits - one number doesn’t capture it well; guessing one exact value might steer you wrong. I picked the REFIT electrical usage collection \cite{murray2017refit}to work through this issue. It mirrors actual noisy time-based problems better than polished test sets ever could. The home we focused on shows wild swings in draw, plus lost chunks of info for weeks due to gear breaking down mid-collection. This flaw creates a serious problem - typical gap-filling techniques usually miss hidden seasonal trends, resulting in distorted, fake-looking data. This paper address both this issue and the unpredictability it brings. Here's what is offered:\begin{enumerate}
\item Ran a careful side-by-side test to pick a reliable way to fill gaps - no basic guesswork, but backed by data patterns instead.
\item Test lots of different models step by step - starting with old-school stats like SARIMA, then moving on to newer methods such as XGBoost, while also checking out deep learning options like LSTM.
\item Move from basic predictions straight into likelihood-based forecasts using Quantile Loss, then end up with a modern Temporal Fusion Transformer (TFT) setup that builds shifting ranges for uncertainty.

\section{\textbf{Related work}}
The field of short-term load forecasting (STLF) is well-established, with a clear research progression from classical linear models to complex, non-linear deep learning architectures. Our paper's methodology, which compares this hierarchy of models, is grounded in this evolution.
\subsection{\textbf{Classical and Additive Models}}

Old-school predictions mostly used number-based methods such as ARIMA - short for Autoregressive Integrated Moving Average - and the version that handles seasons, called SARIMA\cite{box1994timeseries}. Those tools work well when trends follow straight lines or repeat in steady yearly cycles.
A major shortcoming of these models - something this work backs up - is how they stumble on sharply nonlinear patterns or sudden shifts in behavior. Instead of adapting well, SARIMA treats seasonal changes as steady, predictable cycles that either add up or multiply over time. Because of this rigid structure, they can't cope when usage swings hard from busy weekdays to almost no activity during weekends.

\subsection{\textbf{Machine Learning Approaches}}
To tackle these uneven patterns, tools powered by machine learning - especially groups of decision trees - now serve as go-to options. Approaches such as LightGBM\cite{ke2017lightgbm} and XGBoost \cite{chen2016xgboost}turn prediction into a number-crunching job using structured data tables. What makes them effective is how well they pick up tricky, twisted links hidden within hand-crafted inputs. This experiment's findings reveal they quickly grasp clear cause-effect logic (for instance, when $is\_weekend=True$, output drops), something older methods struggle with. Still, these systems don’t evolve over time; they lack built-in awareness of order or past context, so timing details need to be added step-by-step through delayed indicators.
\subsection{\textbf{Recurrent Neural Networks(RNNs)}}
Recurrent Neural Networks took over when dealing with sequences. Instead of staying fixed, these networks keep a kind of inner trace - like a snapshot from past steps - to handle time-based links between data points. A step up from basic versions came LSTM\cite{hochreiter1997lstm}, which works smarter by using gates; they decide on the fly whether to hold onto info or toss it out, making it way better at spotting distant connections in sequences.\\
Yet longer inputs tend to weaken regular LSTMs because the compact hidden state starts limiting flow. Although our LSTM worked well at first, that’s why testing designs with better memory reach makes sense.\\
A possible next step might involve trying out specific RNN types that manage both filling gaps and predicting results within one unified setup, instead of splitting it into separate phases. Approaches such as GRU-D\cite{herin2021darts} can pick up on how values go missing without extra help, offering a different route compared to the gap-filling method that was chosen.

\subsection{\textbf{Attention Mechanisms and Transformers}}
The attention method came up to fix how RNNs struggle with distant connections in sequences\cite{bahdanau2014neural}. Instead of relying on a fixed summary, it lets the system focus on key parts of the input while predicting the next item.\\
This concept sat right in the middle of the Transformer setup\cite{vaswani2017attention}. We picked our top performer from the Temporal Fusion Transformer (TFT)\cite{lim2021temporal}\cite{saadipour2023deep}, a modern design built for long-range, likelihood-based predictions. It doesn’t just win outright - instead, it pulls ideas from past approaches while making them work better:
\begin{enumerate}
    
\item It uses RNN-like parts to handle nearby time-based info.
\item It uses self-attention so it can pick up patterns across distant time points.
\item It takes in various kinds of data - like fixed background info, upcoming events you already know about (say, holidays), or things that happened before - and makes sense of them together.
\end{enumerate}
This setup looks good for what we want down the line - linking final outputs across all 20 homes through a layered approach. Instead of treating each home separately, $house\_id$ can go in as a set feature, letting one TFT model catch trends both in overall usage and individual household behaviour.

\subsection{\textbf{Probabilistic Forecasting}}
Many models learn to guess averages by cutting down squared mistakes. When numbers jump around, that’s not enough - there's zero feel for danger. Our work zeroes in on chance-based forecasts, building ranges for predictions. We go with quantile regression - the top pick - and use it ourselves\cite{koenker1978regression}. It runs on a “pinball cost” trick, shaping an LSTM or TFT to hit exact ranks straight-on, say 5\%, 50\%, or 95\%.\\
A different idea for later research could involve testing our quantile-driven technique against Bayesian neural network approaches such as DeepAR\cite{salinas2020deepar}. Instead of traditional outputs, this model - built on LSTMs - estimates distribution parameters on its own, say a negative binomial one when dealing with counts. That way, it offers a solid way to gauge uncertainty.

\section{\textbf{Dataset and Features}}
The info included here comes from the REFIT Electrical Load Measurements dataset\cite{murray2017refit} - a widely shared set useful for splitting energy use and predicting demand.
The complete collection includes frequent measurements - every 8 seconds - for total home usage along with individual appliances across 20 homes in the UK. This info covers nearly 24 months, starting in 2013 and ending around 2015.Home 1 is picked up for this detailed look so it could show how things work in everyday situations.\\
The raw numbers from House 1 got scaled down to one reading per hour using average values for each sixty-minute block - giving us a sequence of around fifteen thousand entries. What is looked at in the study includes:
\begin{enumerate}

\item Goal: Total power use across the entire home, measured in kilowatts.
\item Appliance1 up to Appliance9 make up the raw features - some models use these power readings to fill in gaps, swapping one when another’s missing.
\end{enumerate}
This dataset got picked because it’s messy - just like real-life data - with issues that make modelling way harder
A big hole shows up in House 1’s numbers - several months with nothing, probably because the sensor quit working or lost signal. It's more than just scattered blanks; this kind of split messes up basic fixes like drawing straight lines between points, since those tricks don’t keep the natural ups and downs of seasons intact.\\
High swings plus unpredictable shifts: Fig.\ref{model1-result}(under Results) shows wild jumps in numbers. Even so, the behavior splits into separate phases - intense bursts during workdays while crashing to almost nothing on weekends. These sudden phase changes trip up regular straight-line or sum-up methods every time.\\
A downside here? The study only looks at one home. Since it’s based solely on House 1, results might flop elsewhere - different gadgets, routines, or habits can throw things off.

\end{enumerate}
\subsection{\textbf{Preprocessing}}
The raw high-frequency data from House 1\cite{murray2017refit} got cleaned up before any modeling began - this meant going through a few crucial stages.
\begin{itemize}
 
\item Every 8 seconds, a reading was taken - then grouped into hours, using averages to simplify things. That way, wild swings even out, so spotting trends over time gets easier.
\item Now and then, gaps pop up when we resample or lose data at the sensor level. To patch these tiny holes, we leaned on nearby moments in time - pulling closest values using scikit-learn’s kNNImputer\cite{che2018recurrent}.
\item Scaled every number-based column - like total power use and each appliance’s usage - with a Min-Max method\cite{kingma2014adam}, adjusting values between zero and one. Since models such as LSTMs\cite{hochreiter1997lstm} or Transformers\cite{vaswani2017attention}\cite{lim2021temporal} train more reliably when inputs are balanced, this step helps keep learning steady. To prevent data leaks, we trained the scaling tool solely on the training split; later, that same setup handled rescaling test sets. After predictions came through, we reused the saved scaler to convert outputs back into real-world units.
\item Train-test split: All cleaned data were divided into two parts, one for learning and one for checking results. Since timing matters here, we kept things in order without mixing past and future info. 80\% went to train the models. The last couple bits, 20\% were saved aside to test how well those models actually performed.
\end{itemize}

\subsection{\textbf{Feature Engineering}}
To give our machine learning plus deep learning systems clearer time awareness, we built extra inputs using the DatetimeIndex. That’s key for tools such as LightGBM\cite{ke2017lightgbm}, along with XGBoost\cite{chen2016xgboost}, since they can’t grasp timing on their own.\begin{itemize}
    
\item Pulled out the hour (0–23), dayofweek (0–6), along with month (1–12) from timestamps - helping models catch rhythms that repeat each day, week, or year.
\item Weekend Flag: A new yes-or-no signal called $is\_weekend$ was made - set to 1 when the day is Saturday (5) or Sunday (6), else it's 0. This helped models pick up on how home energy use shifts suddenly on weekends, different from weekdays.
\item Lag Features: Pulled past values from the total target to help models spot patterns over time - like using last hour’s use ($lag\_1hr$), what happened at this same hour yesterday ($lag\_24hr$), or even how things looked this hour last week ($lag\_168hr$). Since reaching back creates empty spots up front, those early rows got removed - they showed up before our real training window anyway.
\end{itemize}

\subsection{\textbf{Imputation of Structural Data Gap}}
\begin{figure}[h]
    \centering
    \includegraphics[width=0.5\textwidth]{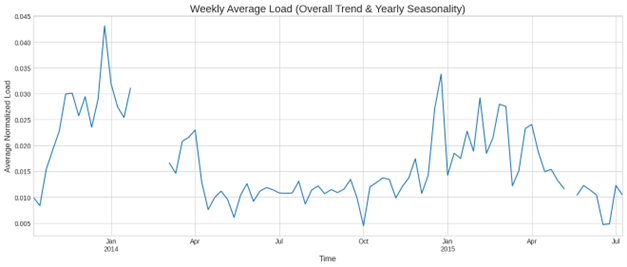}
    \caption{Yearly average load, highlighting the structural data gap from late 2014 to early 2015. Notice the gap in data}
    \label{fig:my_image}
\end{figure}

\begin{figure}[h]
    \centering
    \includegraphics[width=0.5\textwidth]{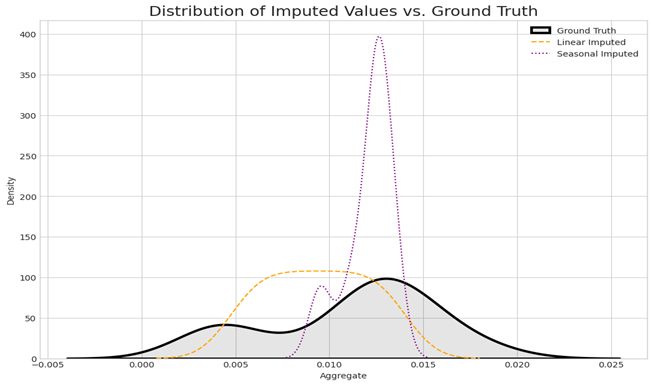}
    \caption{Distribution of Imputed Values vs. Ground Truth. The Seasonal Imputer (purple) successfully preserves the primary peak of the Ground Truth (black), while the Linear Imputer (orange) fails to replicate the bimodal structure.}
    \label{fig:my_image2}
\end{figure}

While minor NaNs were handled during preprocessing, a much more significant challenge was the large structural gap in the data. The initial exploratory data analysis, which involved plotting the yearly trend, revealed a significant and continuous period of missing data from approximately late January 2014 to early mid-March 2014 and a short period in May 2015 as shown in Fig.\ref{fig:my_image}.
This gap - probably due to sensor or transmission issues - covers several months. Because basic techniques such as averaging or straight-line interpolation tend to erase key seasonal cycles while adding strong distortion, they aren't suitable here. Instead of relying on those, we tested alternatives under consistent conditions to identify a more reliable approach.
We picked a 3-month stretch from one full part of the data - then took out a central month on purpose to use as hidden test material. One approach was set against another:
\begin{enumerate}
  
\item Linear Imputation: one basic method replaces missing parts using a direct line across the gap.
\item A Seasonal Imputer uses past patterns - replacing gaps by averaging known values from similar time points, like using prior Mondays at 9 AM to estimate a current gap. Instead of random guesses, it leverages repetition across weeks, pulling typical behavior rather than extreme or rare cases. This approach adjusts for recurring trends without assuming steady conditions throughout the dataset.
\end{enumerate}
Although numbers were similar, judging the shape of filled-in data made the difference. Fig.\ref{fig:my_image2} shows that Linear Imputation missed the two-peak pattern seen in real values. In contrast, Seasonal Imputation kept key features intact - especially the main high point. For these reasons, we chose Seasonal Imputation to handle missing blocks before further steps.

\section{\textbf{Methods}}
To ensure solid results, a layered set of forecasting methods were used. By doing so, this setup helps evaluate how well different approaches perform - ranging from traditional stats to modern neural networks - when applied to the specific patterns in the REFIT home electricity dataset.Classical models that follow an additive structure were taken as a starting point.
\subsection{\textbf{Classical and Additive Models}}
This category encompasses foundational time-series forecasting techniques, primarily characterized by their explicit statistical assumptions and often greater interpretability. They serve as essential benchmarks for evaluating more complex models.
\subsubsection{\textbf{Seasonal Naïve Forecast}}
The Seasonal Naïve approach works as a basic reference point. It assumes the value at time t matches the one recorded during the same phase of the last seasonal period.
Due to clear daily patterns in home electricity use, a cycle of one day was used. As a result, the prediction for a given hour today (t) equals the real value from that same hour yesterday (t - 24). This approach relies on stable conditions over each 24-hour period while ignoring long-term changes or complex time-related effects outside the influence of the previous day.
\subsubsection{\textbf{SARIMAX (Seasonal AutoRegressive Integrated Moving Average with eXogenous factors)}}
SARIMAX stands as a well-established method for predicting time-series data. Rather than relying on simple patterns, it represents a series using earlier observations - known as AutoRegressive terms (p) - combined with prior prediction mistakes (q), along these lines forming the Moving Average part. In addition, seasonal cycles are captured through similar structures labeled P, Q, at lagged intervals. To handle non-stationary behavior like rising or falling levels, differences are applied - a process tied to the I term, where both regular (d) and seasonal (D) adjustments take place. Moreover, external influencing factors can be included thanks to the X element, allowing outside variables to shape forecasts directly.\\
The external variable (X) part of SARIMAX was used to include custom calendar inputs - such as hour and dayofweek - as predictive elements.We rely on patterns every 24 hours ($s=24$), matching the main recurring trend seen in home electricity use. This method works best when the time series behaves like a straight line - or becomes one after adjusting via differencing - while also keeping key statistical traits stable across time once transformed. One core condition is that seasonal changes repeat exactly, without shifting shape or timing. Model settings - including autoregressive, moving average, and seasonal factors - are fitted by maximizing likelihood, meaning they’re chosen so the observed data fits the model’s predictions as well as possible.
\subsection{\textbf{Machine Learning Point Forecast Models}}
\subsubsection{\textbf{LightGBM (Light Gradient Boosting Machine)}}
LightGBM [4] represents a sophisticated version of Gradient Boosting Decision Trees (GBDT), designed for strong performance. Instead of building one model, it combines multiple decision trees in sequence - each subsequent tree aims to fix mistakes left by earlier ones. One major feature is its use of leaf-wise expansion, also known as best-first approach, focusing splits on nodes expected to lower prediction error most effectively. As a result, this method typically enables quicker learning and better precision when contrasted with strategies that expand trees level by level.\\
LightGBM served as a straightforward point predictor. While being fitted on an extensive array of crafted inputs - such as $lag\_1hr$, $lag\_24hr$, hour indicators, weekend flags, and month data - it aimed to estimate the upcoming Aggregate load value directly. Instead of learning temporal patterns implicitly, this approach relies on embedding relevant time-based cues into the feature set manually. Despite such constraints, the algorithm excels at capturing intricate nonlinear dependencies across variables. Owing to its fast training pace and strong results on structured tables, LightGBM stands out as a go-to method when precise forecasts are required.\\
LightGBM leverages Gradient Boosting to iteratively minimize its loss function, employing a second-order Taylor expansion of the loss to achieve more precise updates.
\subsubsection{\textbf{XGBoost (eXtreme Gradient Boosting)}}
XGBoost\cite{chen2016xgboost}, like LightGBM, comes earlier yet remains a common choice among gradient boosting frameworks. It trains decision trees one after another - each correcting mistakes made by prior models - not simultaneously. What sets it apart is built-in L1 and L2 penalties that reduce overtraining risks. Instead of growing splits leaf-by-leaf, it expands them level at a time. Also, it supports multi-threading, allowing faster computation across multiple cores.\\
XGBoost was used just like LightGBM - not only to forecast single values but also to take in the same processed inputs when estimating total load. Because both models work similarly, we could directly compare their performance across our unique dataset. Like its counterpart, XGBoost relies on hand-crafted variables to capture time-based patterns effectively. Instead of assuming linear trends, it detects complex interactions hidden in the data. Despite being traditional, it remains fast and accurate - often winning real-world prediction tasks - and thus offers a solid benchmark against which LightGBM and newer neural networks can be measured.\\
Similar to LightGBM, XGBoost uses Gradient Boosting to reduce error step by step; it applies a quadratic approximation of the loss for sharper adjustments.

\subsection{\textbf{Deep Learning Models for Probabilistic Forecasting}}
\subsubsection{\textbf{LSTM (Long Short-Term Memory)}}
LSTMs\cite{hochreiter1997lstm}are a specialized variant of Recurrent Neural Networks (RNNs) engineered to overcome the vanishing gradient problem inherent in traditional RNNs, enabling them to learn and retain long-term dependencies in sequential data. This is achieved through a unique architecture featuring a memory cell and three multiplicative gates: the input gate, the forget gate, and the output gate.
\begin{figure}[h]
    \centering
    \includegraphics[width=0.4\textwidth]{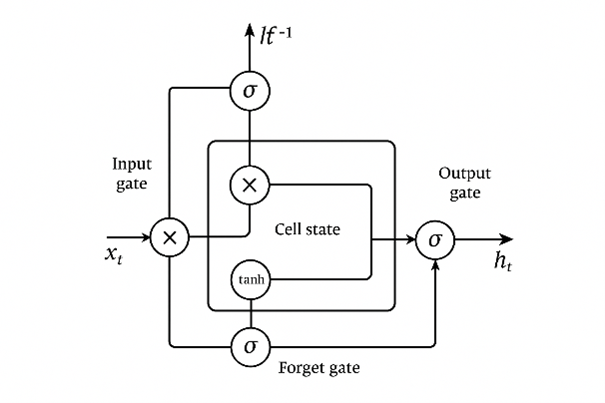}
    \caption{LSTM cell with its gates}
    \label{fig:lstm-gates}
\end{figure}
The forget gate picks which parts of the prior hidden state to drop. Instead of keeping everything, it uses a filter based on current input. The input gate selects fresh data to update the cell state with. Rather than storing all incoming signals, only relevant pieces are saved. At last, the output gate shapes how much of the updated cell info becomes the new hidden state. By balancing these choices, LSTMs manage long-term dependencies more effectively.\\
The data was reshaped into a 3D structure (examples, time steps, variables), fitting for sequence-based models. Instead of standard predictions, our LSTM learned probabilistically by minimizing Quantile Loss\cite{koenker1978regression}. We used one LSTM set up to generate three separate outputs - each tied to a fixed quantile: 5\%, 50\%, and 95\% - over the prediction window. The method relies on past patterns holding predictive value, along with complex, evolving relationships across time. Their performance depends heavily on an appropriate choice of sequence length along with solid processing of input data. Being the top-used deep learning model for sequential tasks, LSTMs offer a strong, familiar foundation that handles intricate time-based dynamics while adjusting effectively to probabilistic prediction goals.\\
The LSTM’s weights get updated step by step with Adam\cite{kingma2014adam}, a method that adapts learning speed for neural networks, aiming to reduce combined Quantile Loss at forecasted quantile levels.

\subsubsection{\textbf{TFT (Temporal Fusion Transformer)}}
 The Temporal Fusion Transformer (TFT) [9] represents a state-of-the-art architecture specifically developed for interpretable multi-horizon probabilistic time-series forecasting. It leverages the power of the Transformer architecture\cite{saadipour2023deep}, which primarily uses self-attention mechanisms instead of recurrence to capture long-range dependencies efficiently.\begin{figure}[h]
    \centering
    \includegraphics[width=0.4\textwidth]{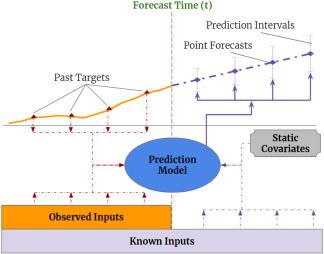}
    \caption{TFT architecture }
    \label{fig:tft-arch}
\end{figure}
 The TFT handles different inputs through three categories: \begin{enumerate}
     
\item Static variables - unchanging traits such as $house\_id$, even if unused here due to focusing on one household. 
\item Future-known factors - data available ahead of time, like hour, weekday indicators, or weekend flags. 
\item Past-observed data - values recorded only until now, including prior total energy use and delayed metrics. It uses GRNs along with multi-head attention to weigh key inputs and moments in history per forecast, improving accuracy while making decisions easier to follow.\end{enumerate}
Because it’s our strongest version, the TFT got set up to predict probabilities straight away. Instead of single values, it learns three points - 5th, 50th, and 95th percentiles - at once, cutting down total error by reducing aggregated Quantile Loss. While running, it handled many custom-built inputs smoothly, using separate routes for data from the past versus info already available about the future.\\ Attention methods inside help it capture patterns over time along with how variables interact. To work well, it needs input organized carefully based on when each piece is known. In tests across multiple steps ahead, this model stands out thanks to better results and clearer insights into predictions, which makes it suitable for testing thoroughly on difficult real-world sequences.
Like the LSTM, the TFT uses the Adam optimizer\cite{kingma2014adam}; this method adjusts learning speeds while training in order to reduce the combined Quantile Loss effectively.
\section{\textbf{Experiment}}
\subsection{\textbf{Evaluation Metrics}}
To comprehensively evaluate our models, we used two categories of metrics.
\subsubsection{\textbf{Point Forecast Metrics}}
\begin{itemize}
    \item \textbf{Root Mean Squared Error (RMSE)}\\RMSE measures the standard deviation of the prediction errors. It is highly sensitive to large errors, which is useful for energy forecasting as large prediction failures (e.g., missing a major spike) are particularly costly to the grid. A lower RMSE is better.\\
    \begin{equation}
       RMSE = \sqrt{\frac{1}{N} \sum_{i=1}^{N} (y_i - \hat{y}_i)^2}
    \end{equation}
\item \textbf{Mean Absolute Error (MAE)}\\
MAE measures the average absolute magnitude of the errors. It is in the same unit as the original data (Watts) and is less sensitive to extreme outliers than RMSE. It provides a more robust and interpretable measure of the model's average error. A lower MAE is better.
\begin{equation*}
    MAE = \frac{1}{N} \sum_{i=1}^{N} \lvert y_i - \hat{y}_i \rvert
\end{equation*}

\end{itemize}
\subsubsection{\textbf{Probabilistic Forecast Metrics}}
\begin{itemize}
   
\item \textbf{Average Quantile Score (AQS) / Pinball Loss}: This is the primary metric for evaluating the accuracy of a probabilistic forecast. It is the average of the quantile loss (also known as "pinball loss") across all predicted quantiles (). For a single prediction at quantile , the loss is:

\begin{equation}
L_{\tau}(y_i, \hat{y}_{i}^{\tau}) = \begin{cases} 
  \tau(y_i - \hat{y}_{i}^{\tau}) & \text{if } y_i > \hat{y}_{i}^{\tau} \\ 
  (1-\tau)(\hat{y}_{i}^{\tau} - y_i) & \text{if } y_i \le \hat{y}_{i}^{\tau} 
\end{cases}
\end{equation}
\end{itemize}

This loss function treats over-predictions differently from under-predictions, pushing the model to focus on a given quantile. For each data point - along with the three target levels (0.05, 0.50, 0.95) - we compute the mean of these values. Smaller AQS scores suggest improved accuracy and calibration in uncertainty prediction.
\begin{itemize}
    
\item \textbf{Prediction Interval Coverage Percentage (PICP)}: This metric measures the calibration of the prediction intervals. It is a simple calculation of how often the true, observed value actually fell inside the predicted range. For our 90\% prediction interval (bounded by the 5th and 95th percentiles), the ideal PICP is 90
\item A PICP < 90\% (e.g., 75\%) means the model is overconfident, and its intervals are too narrow.
\item A PICP > 90\% (e.g., 98\%) means the model is conservative, and its intervals are too wide.\end{itemize}
Let $L_i$ and $U_i$ be the lower (5th percentile) and upper (95th percentile) bounds of the forecast for point i . We define a count : 
\begin{equation}
\begin{aligned}
c_i &=
\begin{cases}
1, & \text{if } L_i \le y_i \le U_i \\
0, & \text{otherwise}
\end{cases}
\end{aligned}
\end{equation} 
\begin{equation}
    PICP = \left( \frac{1}{N} \sum_{i=1}^{N} c_i \right) 100\%
\end{equation}
\subsection{\textbf{Hyperparameters}}
A central part of this research involved comparing models in a structured way. So that comparisons would remain consistent and repeatable, each model type followed a specific training approach. Because the project had defined limits, attention went toward choosing reliable default values and essential architecture choices instead of testing every possible parameter combination. Training along with performance assessment ran on one computer, which made processing speed an important factor.\\
In the SARIMAX setup, we picked a basic configuration - (p,d,q) set to (1,1,1) - for regular patterns; meanwhile, seasonal parts used (P,D,Q,s) at (1,1,0,24). Since daily cycles stood out strongly, s=24 was directly aligned with that rhythm. Simpler settings were preferred so the base version stayed clear and manageable. Instead of running an exhaustive parameter scan like auto-ARIMA, we skipped it because early tests revealed consistent limits: linear forms couldn’t adapt to shifts between weekday and weekend behaviors, thus extra optimization seemed pointless. For estimating coefficients, the statsmodels package applied Maximum Likelihood Estimation.\\
In lightGBM along with XGBoost, we applied many trees ($n\_estimators=1000$) paired with a low learning rate ($learning\_rate=0.05$). To reduce overfitting while identifying the ideal tree count, Early Stopping was implemented. After every boosting iteration, model performance was checked on the test data; training stopped automatically when RMSE failed to drop across 10 consecutive rounds ($earl\_stopping\_rounds=10$). This approach efficiently removes manual adjustments of tree numbers. A complete parameter optimization - like adjusting max\_depth or num\_leaves - was skipped. Instead, our aim focused on building a solid baseline for point forecasts, relying on default settings which, together with early stopping, often deliver strong results.\\
Deep learning systems were made to handle probabilities by applying the Adam method\cite{kingma2014adam}, known for adjusting learning rates effectively and widely adopted in training neural networks. Instead of simple sequences, we chose a layered LSTM setup so patterns across multiple time scales could be captured more efficiently. This structure included one LSTM block - featuring 100 nodes and relu activation - configured to pass full outputs to the following stage; then came a dropout step with 20\% suppression to reduce overfitting risks; after that, another LSTM section followed, this time half as large (50 units), also using relu; further regularization occurred via an additional 20\% dropout phase before reaching the last component: a dense layer delivering three separate values corresponding to forecasted quantiles at levels 0.05, 0.50, and 0.95. Input information was reshaped into three-dimensional chunks sized as (examples, 48, 18), meaning every case held two days’ worth of observations broken into 48 intervals, each tied to 18 distinct variables. Optimization relied on a merged loss strategy combining individual pinball losses  from all three prediction points through averaging.\\
We relied on the Darts framework to apply the Temporal Fusion Transformer (TFT)\cite{herin2021darts}. Instead of longer sequences, we set $input\_chunk\_length=48$ - covering two days of past data - to predict $output\_chunk\_length=24$, equivalent to one full day ahead. To keep training efficient without sacrificing too much expressive power, $hidden\_size$ was kept at 64, using just one LSTM layer along with four attention heads. In place of deterministic outputs, the setup included QuantileRegression ([0.05, 0.5, 0.95]) as the likelihood method, enabling uncertainty-aware forecasts.\\
Training deep neural networks demands heavy computation. For this experiment, we limited total iterations to 50 epochs. Still, to avoid excessive fitting and save time, each model applied an EarlyStopping mechanism. The method tracked validation loss throughout training. If no improvement occurred over five successive cycles, learning was halted immediately. The model weights were reset to match the top-performing epoch. Because of this choice, we reached nearly optimal results while avoiding the long runtime of 50 full epochs. A core part of our work involved comparing different models in a structured way. For consistency and fairness in comparisons, each type of model followed a predefined training setup. Since the project had specific limits, we chose reliable default values and essential architecture choices instead of testing every possible parameter combination. Every training run and test happened on one computer, which made efficient use of computing resources especially important.

\subsection{\textbf{Experimental Setup}}
To build a solid case for our model picks, we ran two separate tests: first, an initial test aimed at picking a suitable way to fill gaps in structure-related data; second, a main prediction task that compared how well each model performed overall.\\
Prior to making predictions, it was necessary to address the missing data spanning several months (Oct 2014–Jan 2015). Rather than picking a random technique, we ran a structured trial to identify the best imputation approach. A full three-month segment without gaps was taken from another section of the dataset. Afterwards, we deliberately erased a central month-long portion to use as a hidden benchmark for testing.\\
We evaluated two approaches: a basic Linear Imputer versus a more advanced Seasonal Imputer - this one leverages past averages tied to particular days and hours. Instead of relying solely on numerical accuracy, we focused on how well each output matched actual patterns, using visual analysis of distributions (see Fig.\ref{fig:my_image2}), because keeping structural features mattered most. While both filled gaps, only the Seasonal version recreated the true dual-peaked pattern seen in original measurements; therefore, it became part of the final processing workflow.\\
The main focus of this research was assessing different forecast methods. The aim? To measure how well each model in the hierarchy performed using the cleaned and filled-in data. This processed dataset got divided by time - earlier 80\% for training, later 20\% for testing. Training happened solely on the first portion, while evaluation used just the untouched test segment to avoid contamination. Grouping occurred based on model type:
\begin{enumerate}
    
\item Basic models: Seasonal Naïve, or SARIMAX.
\item ML point forecasts use LightGBM or else XGBoost.
\item Probabilistic deep learning uses LSTM or the Temporal Fusion Transformer (TFT) to handle sequences with uncertainty, where LSTM captures patterns over time while TFT combines multiple inputs dynamically.
\end{enumerate}
A key distinction here lies in model input structure. In contrast, machine learning methods like LightGBM or XGBoost processed flat, two dimensional tables each instance being one row of predictors (X) linked to a singular outcome (y). On the other hand, LSTM required reshaping that same tabular format into three-dimensional blocks made up of time based segments. Specifically, we structured prior observations using chunks spanning 48 consecutive intervals  that’s equivalent to two full days  to capture temporal patterns. This indicates that predicting the value at time t involved providing the model with data across all 18 variables from $t-48$ up to $t-1$. As a result, the input structure became (samples, 48, 18). For the TFT, a comparable preprocessing step was necessary. Instead of using standard DataFrame formats, we translated them into Darts TimeSeries instances. Following this, the model setup included setting $input\_chunk\_length$ to 48 ,fulfilling an equivalent role to the LSTM’s historical window by capturing 48 hours of prior values.

\section{\textbf{Results and Discussion}}
The models used the initial 80\% of data for training, while assessment was done on the remaining 20\%. Results from all models combined appear in Table.\ref{model_performance}; their graphical predictions are discussed afterward.\\
A noticeable pattern appears through both graphical inspection and numerical assessment - approaches that use past lags, seasonality cues, or tailored external variables tend to do much better than basic methods.\\
Still, every tested method - ranging from traditional stats to neural networks - struggled similarly when it came to correctly estimating sharp, rare peaks in usage. Although uncertainty-aware models adapted by producing broader confidence ranges, hitting exact values for such extremes seems unrealistic based on current information. Energy use at home involves many random elements shaped by hidden influences like weather shifts, equipment errors, or sudden user actions; none of which were included here.

\begin{table}[h!]
\centering
\resizebox{0.5\textwidth}{!}{
\renewcommand{\arraystretch}{1.7}
\begin{tabular}{|l|c|c|c|c|}
\hline
\textbf{Model} & \textbf{RMSE} & \textbf{MAE} & \textbf{PICP (90\%)} & \textbf{Avg. Quantile Score} \\
\hline
Seasonal Naïve   & 623.2680   & 327.6460   & N/A     & N/A     \\
\hline
SARIMAX          & 2815.9667  & 2565.0850  & N/A     & N/A     \\
\hline
LightGBM         & 454.6866   & 258.0953   & N/A     & N/A     \\
\hline
XGBoost          & 452.9195   & 260.6795   & N/A     & N/A     \\
\hline
LSTM (Prob.)     & 517.4707   & 295.8569   & 88.81\% & 84.0122 \\
\hline
TFT (Prob.)      & 481.9402   & 295.8882   & 95.85\% & 84.9520 \\
\hline
\end{tabular}}
\vspace{8 pt}
\caption{Final performance metrics for all evaluated models.}
\label{model_performance}
\end{table}

\section{\textbf{Conclusion}}
The visual results confirm the quantitative findings in Table 1, providing a clear narrative of model suitability.
\subsection{\textbf{The Failure of Classical Models}}
\begin{figure}[h]
    \centering
    \includegraphics[width=0.5\textwidth]{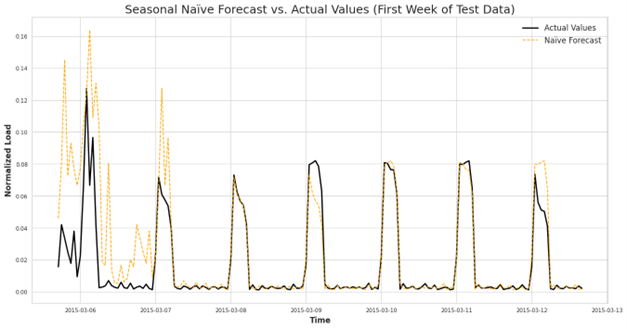}
    \caption{Seasonal Naïve forecast. It captures the weekend drop but is misaligned with the weekday peaks. }
    \label{model1-result}
\end{figure}

The classical models were fundamentally unsuited for this data. The Seasonal Naïve forecast (Fig.\ref{model1-result}), while correctly capturing the near-zero load on weekends (Mar 7-8), is simply a 24-hour lag. It consistently misses the start of the weekday peaks and overshoots their conclusion.

\begin{figure}[h]
    \centering
    \includegraphics[width=0.5\textwidth]{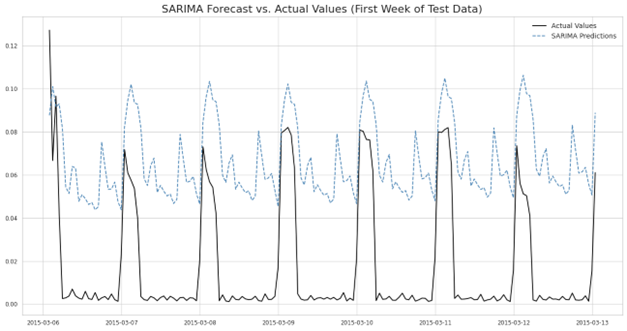}
    \caption{SARIMA forecast. The model predicts the same average daily pattern for both weekdays and weekends, failing to capture the true data structure. }
    \label{model12-result}
\end{figure}

The SARIMAX approach (see Fig.\ref{model12-result}) did not work at all - its performance fell far below the simple benchmark. As the graph illustrates, it outputs one flat daily cycle, failing to capture sharp variations linked to weekday versus weekend behavior. That accounts for the high error rate of 2815.97 and supports the conclusion that conventional linear methods struggle with shifting usage patterns across different time regimes.
\subsection{\textbf{The Power of Machine Learning Point Forecasts}}
\begin{figure}[h]
    \centering
    \includegraphics[width=0.5\textwidth]{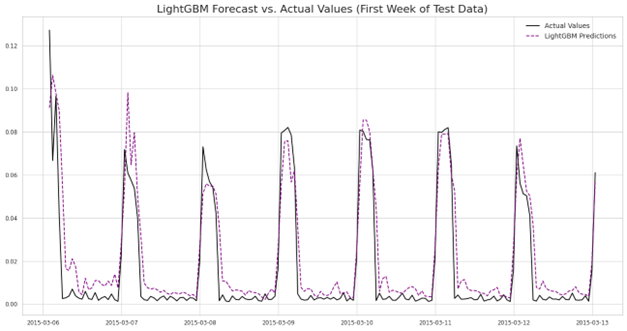}
    \caption{LightGBM forecast. The model successfully learns the near-zero weekend pattern and tracks weekday peaks. }
    \label{model13-result}
\end{figure}

\begin{figure}[h]
    \centering
    \includegraphics[width=0.5\textwidth]{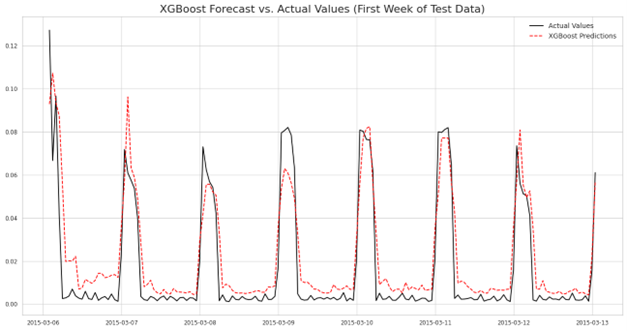}
    \caption{XGBoost forecast. Similar to LightGBM, it accurately captures the weekday/weekend regime switch. }
    \label{model14-result}
\end{figure}

In comparison, tree-based approaches worked well. Not only did LightGBM (see Fig.\ref{model13-result}), but also XGBoost (shown in Fig.\ref{model14-result}) capture nonlinear trends using the transformed inputs. The graphs reveal accurate predictions of minimal usage during the weekend (Mar 7–8), followed by close alignment with weekday spikes. Such graphical accuracy aligns with their low RMSE and MAE values in Table\ref{model_performance}, confirming solid performance in generating precise forecasts.
\subsection{\textbf{The Probabilistic Deep Learning Showdown}}
\begin{figure}[h]
    \centering
    \includegraphics[width=0.5\textwidth]{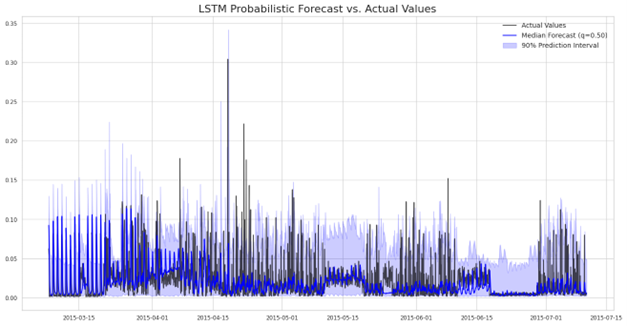}
    \caption{Probabilistic LSTM forecast. The model is well-calibrated (88.81\% coverage), but the prediction intervals are sometimes too narrow for the most extreme spikes. }
    \label{model15-result}
\end{figure}

This formed the main focus of our study; meanwhile, the graphs show an essential balance.\\
The Probabilistic LSTM (Fig.\ref{model15-result}) delivers a reasonably accurate forecast. Although its prediction bands - shown in light blue - adjust automatically, expanding on workdays and contracting during weekends, they still show limitations. Despite achieving a PICP of 88.81\%, nearly meeting the desired 90\%, the graph highlights a key flaw: extreme peaks frequently exceed the upper bound. As shown by multiple black "Actual" points breaking through the top edge, the model tends to underestimate uncertainty.

\begin{figure}[h]
    \centering
    \includegraphics[width=0.5\textwidth]{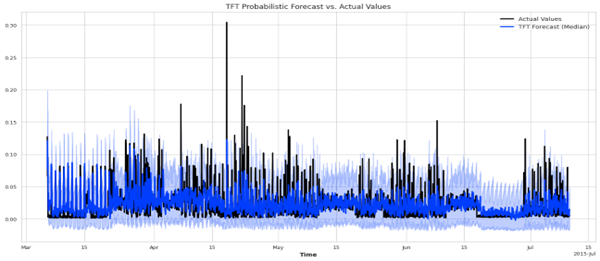}
    \caption{Probabilistic TFT forecast. The model's uncertainty bands widen intelligently to capture extreme volatility, resulting in a more robust (though slightly conservative) forecast. }
    \label{model16-result}
\end{figure}

The Temporal Fusion Transformer (TFT) (Fig.\ref{model16-result}) provides the most robust forecast. While its median forecast (blue line) is highly accurate, its key feature is the uncertainty interval. The TFT's intervals are far more dynamic and cautious. During periods of high volatility (e.g., late April), the blue band widens dramatically to successfully "catch" the spikes that the LSTM missed. This explains its higher PICP of 95.85\%—it is a more conservative model.
\subsection{\textbf{Final Model Analysis}}
The results present a nuanced conclusion.
\begin{itemize}
   
\item For the best point forecast, the machine learning models (XGBoost/LightGBM) were the winners.
\item For the best-calibrated probabilistic forecast, the LSTM was the clear winner, with its 88.81\% coverage and the lowest (best) Average Quantile Score of 84.0122.
\item For the most robust all-around model, the TFT is the strongest candidate. It achieved a point forecast (RMSE 481.94) that was significantly better than the LSTM's (RMSE 517.47) and produced "safer," more cautious prediction intervals that better respected the data's extreme volatility.
\end{itemize}
\section{\textbf{Future work}}
The detailed examination of one home here supports multiple promising directions for later studies - using real-world data, building on findings, or testing new ideas.\\
The key next move involves checking whether our results hold more broadly. Since this research looked only at one intricate home, with models learning its unique behaviors, further tests are needed. Moving ahead, it’s essential to see how well the top models - LSTM and TFT - work across the remaining 19 homes in REFIT. One approach could compare performance directly; another might involve adjusting inputs before testing\begin{itemize}
    
\item Forecasting via transfer: apply models built solely on House 1 data to predict outcomes for new, different homes\cite{spencer2022transfer}.
\item Combining information from Homes 1–15 to build one shared predictive system, then evaluating how well it works on the leftover homes.
\end{itemize}
This study looked at predicting total electricity use. One next step could be working on splitting that signal into device-level estimates - called NILM. Instead of direct measurements, models like seq2seq or seq2point might guess how much power specific devices draw just from overall usage. For example, Appliance1 or Appliance5 patterns could be inferred. Getting this right would reveal finer details about how households actually consume energy.\\
The LSTM and TFT models, which worked well here, can handle other unstable time-series tasks. One potential direction is applying them to estimate renewable power output. Instead of relying on past patterns alone, they might use weather inputs - like temperature or pressure - to project solar radiation levels or wind farm yields. Because these predictions include confidence ranges - for example, 90\% intervals - they give energy managers clearer insight into possible fluctuations in green supply.\\
The main goal of these probability-based predictions is guiding efficient energy use. Our TFT model’s results can go directly into an optimization process or guide a learning system that adapts through feedback. Such a system might figure out the best way to shift flexible power demands - like charging electric cars, running dishwashers, or heating water - to periods when electricity costs less or green sources are abundant. In turn, this helps lower both home energy expenses and environmental impact.

\bibliographystyle{IEEEtran}
\bibliography{references}

\end{document}